\documentclass{article} 
\usepackage{iclr2025_conference,times}

\usepackage{amsmath,amsfonts,bm}

\def\eqref#1{equation~\ref{#1}}

\def\1{\bm{1}}

\DeclareMathAlphabet{\mathsfit}{\encodingdefault}{\sfdefault}{m}{sl}
\SetMathAlphabet{\mathsfit}{bold}{\encodingdefault}{\sfdefault}{bx}{n}

\DeclareMathOperator*{\argmax}{arg\,max}
\DeclareMathOperator*{\argmin}{arg\,min}

\usepackage{hyperref}
\usepackage{url}
\usepackage{amsmath}
\usepackage{amssymb}
\usepackage{mathtools}
\usepackage{booktabs}
\usepackage{multirow}
\usepackage{amsthm}
\usepackage{lipsum}
\usepackage{microtype}
\usepackage{graphicx}
\usepackage{subfigure}
\usepackage{enumitem}
\usepackage{hyperref}
\usepackage{wrapfig}
\usepackage{algorithm}
\usepackage{algpseudocode}
\theoremstyle{plain}
\newtheorem{theorem}{Theorem}[section]

\newtheorem{corollary}[theorem]{Corollary}
\theoremstyle{definition}
\newtheorem{definition}[theorem]{Definition}

\theoremstyle{remark}

\title{Why Uncertainty Calibration Matters for Reliable Perturbation-based Explanations}

\iclrfinalcopy
\author{Thomas Decker$^{1,2,3}$, Volker Tresp$^{2,3}$ \& Florian Buettner$^{4,5,6}$ \\
$^1$Siemens AG \quad  $^2$LMU Munich \quad 
$^3$Munich Center for Machine Learning (MCML) \\ $^4$Goethe University Frankfurt \quad $^5$German Cancer Research Center (DKFZ)\\
$^6$German Cancer Consortium (DKTK) \\
\texttt{thomas.decker@siemens.com} \quad \texttt{volker.tresp@lmu.de} \\ \texttt{florian.buettner@dkfz.de}
}

\begin{document}

\maketitle

\begin{abstract}
Perturbation-based explanations are widely utilized to enhance the transparency of modern machine-learning models. However, their reliability is often compromised by the unknown model behavior under the specific perturbations used. This paper investigates the relationship between uncertainty calibration - the alignment of model confidence with actual accuracy - and perturbation-based explanations. We show that models frequently produce unreliable probability estimates when subjected to explainability-specific perturbations and theoretically prove that this directly undermines explanation quality. To address this, we introduce ReCalX, a novel approach to recalibrate models for improved perturbation-based explanations while preserving their original predictions. Experiments on popular computer vision models demonstrate that our calibration strategy produces explanations that are more aligned with human perception and actual object locations.
\end{abstract}

\section{Introduction}
The ability to explain model decisions and ensure accurate confidence estimates are fundamental requirements for deploying machine learning systems responsibly \citep{kaur2022trustworthy}. Perturbation-based techniques \citep{robnik2018perturbation} have been established as a popular way to enhance model transparency in practice \citep{bhatt2020explainable}. Such methods systematically modify input features to quantify their importance by evaluating and aggregating subsequent changes in model outputs \citep{covert2021explaining}. This intuitive principle and the flexibility to explain any prediction in a model-agnostic way has led to widespread adoption across various domains \citep{ivanovs2021perturbation}.

Nevertheless, the application of perturbation-based techniques faces a fundamental challenge: These methods operate by generating inputs that differ substantially from the training distribution \citep{hase2021out,frye2021shapley} and models often produce invalid outputs for such perturbed samples \citep{feng2018pathologies, carter2021overinterpretation, jain2022missingness}. 
Consequently, forming explanations by aggregating misleading predictions under perturbations can significantly distort the outcome and compromise its fidelity. This is particularly concerning when explanations are used to gain new scientific knowledge through the lens of a machine learning model, where faithful interpretations are essential. 

These issues naturally raise the question of how to attain reliable model outputs under the specific perturbations used when deriving explanations. A classical approach for this purpose is uncertainty calibration \citep{niculescu2005predicting, naeini2015obtaining, guo2017calibration}. It aims to ensure that a model's confidence aligns with its actual accuracy, which is crucial to obtaining meaningful and reliable probabilistic predictions. While calibration has been extensively studied in the machine learning literature \citep{wang2023calibration}, its role in explanation methods remains largely unexplored. First empirical evidence suggests that basic calibration might indeed benefit explainability \citep{scafarto2022calibrate, lofstrom2024calibrated}, but a rigorous theoretical understanding is still missing. 
In this work, we provide the first comprehensive analysis of the relationship between uncertainty calibration and perturbation-based explanations. Our findings establish calibration as a fundamental prerequisite for reliable model explanations and propose a practical solution for enhancing the trustworthiness of perturbation-based explanation methods.

More precisely, we make the following contributions:
\begin{itemize}
    \item We provide a rigorous theoretical analysis revealing how poor calibration can affect the results of common perturbation-based explainability techniques.
    \item We propose, ReCalX, a novel approach that increases the reliability of model outputs under the particular perturbations used to derive explanations.
    \item We empirically validate the efficacy of ReCalX on a variety of popular image classifiers and demonstrate that the corresponding explanations are more aligned with human perception and actual object positions.
\end{itemize}

\section{Background and Related Work}
\paragraph{Basic Notation}
Consider a probability space $ (\Omega, \mathcal{F}, P) $, where $ \Omega $ is the sample space, $ \mathcal{F} $ is a $\sigma$-algebra of events, and $ P $ is a probability measure. Let $ X: \Omega \to \mathcal{X} $ be a random variable representing the feature space $ \mathcal{X} $, and let $ Y: \Omega \to \mathcal{Y} $ be a random variable representing the target space $ \mathcal{Y} $. The join data distribution of $(X, Y)$ is denoted by $P_{X,Y}$, the conditional distribution of $ Y $ given $ X $ by $ P_{Y | X} $, and the marginal distributions by $P_X, P_Y $. During our theoretical analysis, we will also make use of the following quantities. The \textit{mutual information} between two random variables $ X $ and $ Y $, measures the reduction in uncertainty of one variable given the other and is defined as:
\begin{align*}
I(X, Y) := \mathbb{E}_{(X, Y)} \left[ \log \frac{P_{X,Y}(X,Y)}{P_X(X) P_Y(Y)} \right].
\end{align*}
A related measure is the \textit{Kullback-Leibler (KL) divergence}, which expresses the difference between two probability distributions $ P $ and $ Q $ over the same sample space:
\begin{align*}
D_{\text{KL}}(P \| Q) := \mathbb{E}_{X} \left[ \log \frac{P(X)}{Q(X)} \right].
\end{align*}

\paragraph{Perturbation-based Explanations}
Perturbation-based explanations quantify the importance of individual input features by evaluating how a model output changes under specific input corruptions. Let $ f: \mathcal{X} \to [0,1]^K $ be a classification model over $K$ classes, where $\mathcal{X} \subseteq \mathbb{R}^d$ is the $d$-dimensional input space. Given a sample $x \in \mathcal{X}$, a perturbation-based explanation produces an importance vector $\phi(x) \in \mathbb{R}^d$, where $\phi_i(x)$ quantifies the importance of feature $x_i$. To derive $\phi(x)$, we first introduce a perturbation function $ \pi: \mathcal{X} \times 2^{\{1,\dots, d\}} \to \mathcal{X} $ that, given an instance $ x $ and a subset of feature indices $ S \subseteq \{1, \dots, d\} $, returns a perturbed instance $ \pi(x, S) $ where features in $ S $ remain unchanged while the complementary features $ \bar{S} = \{1,\dots,d\} \setminus S $ are modified. Various perturbation strategies have been investigated \citep{sundararajan2020many, covert2021explaining}, with a common approach being the replacement of features with fixed baseline values such as zeros \citep{sturmfels2020visualizing, jain2022missingness}. The model prediction under such perturbation is given by
:\begin{align*} f^{\pi}_S(x) := f\bigl( \pi(x, S) \bigr). \end{align*}

To compute the final explanation $\phi$, the outcomes of $f_S$ for different subsets $S$ are systematically examined and aggregated. While numerous aggregation strategies exist \citep{covert2021explaining, petsiuk2018rise, fel2021look, novello2022making, zeiler2014visualizing, ribeiro2016should, lundberg2017unified}, we focus on methods that employ a linear summary strategy \citep{lin2024robustness}. In this case, there exists a summary matrix $A \in \mathbb{R}^{d \times 2^d}$ such that $\phi(x) = A\vartheta(x)$, where $\vartheta \in \mathbb{R}^{2^d}$ contains the predictions $f_S(x)$ for all possible subset perturbations $S$. Two prominent methods in this category are Shapley Values \citep{lundberg2017unified} and LIME \citep{ribeiro2016should}.

\paragraph{Uncertainty Calibration}
In classification, a model is said to be \textit{calibrated} if its predicted confidence levels correspond to empirical accuracies, for all confidence levels. Formally, for a classifier $ f: \mathcal{X} \to [0,1]^K $ calibration requires that $P_{Y|f(X)} = f(X)$ \citep{gruber2022better}. Intuitively, this also implies, that for any confidence level $ p \in [0,1] $ it holds :
\begin{align*}
 P(Y = k \mid f(X)_k = p) = p \quad \forall k \in \{1,\dots,K\},
\end{align*}
where $ f(X)_k $ denotes the predicted probability for class $ k $ and $ Y $ is the true label. This means that among all instances for which the model predicts a class with probability $ p $, approximately a $ p $ fraction should be correctly classified. In practice, modern neural classifiers often exhibit miscalibration \citep{guo2017calibration, minderer2021revisiting}, in particular when facing out-of-distribution samples \citep{ovadia2019can, tomani2021post, yu2022robust}. This can be formalized using a \textit{calibration error (CE)}, which quantifies the mismatch between $f(X)$ and $P_{Y|f(X)}$.
In this work, we will consider the KL-Divergence-based calibration error \citep{popordanoska2024consistent}, as this is the proper calibration error directly induced by the cross-entropy loss which is typically optimized for in classification:
\begin{align*}
    CE_{\textit{KL}}(f) = \mathbb{E}\Big[D_{\textit{KL}}\Big(P_{Y|f(X)} \| f(X)\Big)\Big]
\end{align*}
Our work investigates how violations of this calibration property affect the quality of the derived explanations $\phi(x)$ and develops methods to improve calibration specifically for perturbation-based techniques, which has not been analyzed before.
\section{Understanding the Effect of Uncertainty Calibration on Explanations}
In this section, we provide a thorough mathematical analysis how poor calibration under explainability-specific perturbation can degrade global and local explanation results. All underlying proofs and further mathematical details are provided in the appendix. 

\subsection{Effect on Global Explanations}
Calibration is inherently a global property of a model related to the entire data distribution and in general, it is impossible to evaluate it for individual samples \citep{zhao2020individual, luo2022local}. Hence, we first analyze its effect on global explanations about the model behavior \citep{lundberg2020local} such as the importance of individual features for overall model performance \citep{covert2020understanding, decker2024explanatory}. Following \citep{covert2020understanding}, we define the global predictive power of a feature subset $S$ for a model $f$, as follows:
\begin{definition}
    Let $S\subset \{1, \dots, d\}$ be a feature subset and let $f_S^{\pi}$ be the model that only observes features in $S$ while the other are perturbed using a predefined strategy $\pi$, then the predictive power $v_f^{\pi}$ (S) for a model $f$ and a loss function $\mathcal{L}: \mathcal{Y} \times \mathcal{Y}\to \mathbb{R}^+$ is defined as:
    \begin{align*}
        v_{f}^{\pi}(S) = \mathbb{E}[\mathcal{L}(f_{\emptyset}^{\pi}(X), Y) ] - \mathbb{E}[\mathcal{L}(f_{S}^{\pi}(X), Y) ]
    \end{align*}
\end{definition}
Note, that the predictive power captures the performance increase resulting from observing features in $S$ compared to a baseline prediction where all features are perturbed using $\pi$. In classification, models are typically optimized for the cross-entropy loss $\mathcal{L}_{\textit{CE}}(f(x),y) = -\log(f(x)_y)$.  We show that in this case the predictive power can be decomposed as follows:
\begin{theorem}
    Let $\mathcal{L}_{\textit{CE}}$ be the cross-entropy loss, $D_{\textit{KL}}(\cdot,\cdot)$ be the KL-Divergence between two distributions, and let $I(\cdot,\cdot)$ denote the mutual information between random variables. Then we have:
    \begin{align*}
    v_{f}^{\pi}(S) &= \underbrace{D_{\textit{KL}}(P_Y\|f_{\emptyset}^{\pi}(X))}_{\text{Perturbation Baseline Bias}} + 
    \underbrace{I(f_S^{\pi}(X),Y)}_{\text{Information in $f_S^{\pi}$ about $Y$}} - 
    \underbrace{CE_{\textit{KL}}(f_S^{\pi})}_{\text{Calibration Error of $f_S^{\pi}$}}
    \end{align*}
\end{theorem}
This theorem is in line with existing decomposition results \citep{brocker2009reliability,kull2015novel, gruber2022better} that hold for a general class of performance measures related to proper scoring rules \citep{gneiting2007strictly}, including $\mathcal{L}_{\textit{CE}}$. It further allows for an intuitive interpretation of what drives the predictive power when evaluated with a specific perturbation and reveals a fundamental relationship with calibration. The first term can be interpreted as bias due to the perturbation strategy, implying that an optimal uninformative baseline prediction should yield $P_Y$. The second term is the mutual information between the target $Y$ and the model that only observes features in $S$. This corresponds to the reduction in uncertainty about the target $Y$ that knowledge about $f_S^{\pi}$ provides. The third term is the KL-Divergence-based calibration error of the restricted model $f_S^{\pi}$ when facing the perturbation $\pi$. The theorem implies that if the model to be explained produces unreliable predictions under the perturbation used, then the resulting calibration error directly undermines the predictive power. Moreover, Theorem 3.2 has the following immediate consequence:
\begin{corollary}
    If a model $f$ is perfectly calibrated under all subset perturbations faced during the explanation process, then we have:
    \begin{align*}
        v_{f}^{\pi}(S) = I(f_S^{\pi}(X), Y)
    \end{align*}
\end{corollary}
This implies that $I(f_S^{\pi}(X),Y)$ can be considered as an idealized predictive power that can only be attained when the model always gives perfectly calibrated predictions under every perturbation faced. Note this also directly extends a corresponding result in \citep{covert2020understanding}, showing that this relationship holds for the naive Bayes classifier $P_{Y|X}$. It rather holds for any model that is perfectly calibrated under the perturbations faced, which is also the case for $P_{Y|X}$. Moreover, explaining by directly approximating similar mutual information objectives has also been proposed as a distinct way to enhance model transparency \citep{chen2018learning, schulz2020restricting}.

\subsection{Effect on local explanations}
Local explanations enable the understanding of which features are important for an individual prediction. As emphasized above, calibration is a group-level property so evaluating it for a single sample is impossible. Nevertheless, we can still estimate how an overall calibration error under the perturbations might propagate to an individual explanation result:

\begin{theorem} 
Let $\phi(x)$ be a local explanation for input $x$ with respect to model prediction $f(x)$ and perturbation $\pi$. Let $\phi^{\ast}(x)$ denote the explanation that would be obtained if the model $f$ were perfectly calibrated under all subset perturbations. Define the maximum calibration error across all perturbation subsets as: $ CE_{\textit{KL}}^{\max_S}:= \max_{S\subset [d]}CE_{\textit{KL}}(f_S^{\pi}) $ Then, with probability at least $(1-\delta)$, the mean squared difference between the actual and ideal explanations is bounded by:
\begin{align*} 
\frac{1}{d}\lVert \phi(x) - \phi^{\ast}(x) \rVert_2^2 \le 2 CE_{\textit{KL}}^{\max_S} + \sqrt{8\log(1/\delta)}
\end{align*} 
\end{theorem}
The theorem shows that poor calibration under perturbations can at worst also directly impact the quality of a local explanation result $\phi(x)$. Importantly, it demonstrates that to improve explanation quality through uncertainty calibration, it is insufficient to only reduce the calibration error on unperturbed samples, as proposed by \citep{scafarto2022calibrate}. Instead, reliable explanations explicitly require models to be well-calibrated under all subset perturbations $\pi(\cdot, S)$ used in the explanation process. In the next section, we propose an actionable technique, ReCalX, to achieve this goal. 

\section{Improving Explanations with Perturbation-specific Recalibration (ReCalX)} 
Recalibration techniques aim to improve the reliability of the probability estimates for an already trained model. These post-hoc methods transform the model's original outputs to better align predicted probabilities with empirical frequencies. Among various approaches \citep{silva2023classifier}, temperature scaling has emerged as the most widely adopted method due to its simplicity and empirical effectiveness \citep{guo2017calibration, minderer2021revisiting}.
Formally, given a model $f: \mathcal{X} \rightarrow [0,1]^K$ that outputs logits $z(x) \in \mathbb{R}^K$ before the final softmax layer, temperature scaling introduces a single scalar parameter $T > 0$ that divides the logits before applying the softmax function:
\begin{align*} f(x; T)_k := \frac{\exp(z_k(x)/T)}{\sum_{j=1}^K \exp(z_j(x)/T)} \end{align*}
The temperature parameter $T$ is optimized on a held-out validation set,
typically by minimizing the cross entropy loss $\mathcal{L}_{\textit{CE}}$ in the case of classification.

Importantly, temperature scaling preserves the model's predicted class as $\argmax_k f(x; T)_k = \argmax_k f(x)_k$ for all $T>0$ \citep{zhang2020mix}. This property is particularly important when aiming to improve explanation as it also ensures that recalibration will not change the underlying model behavior that we seek to explain. 

Going beyond classical temperature scaling we propose $\text{ReCalX}_{\textit{TS}}$ as an augmented version designed to enhance perturbation-based explanation. It aims to reduce the calibration error under all perturbations faced during the explanation process by scaling logits using an adaptive temperature that depends on the perturbation level. 
Formally, for a subset $S \subseteq \{1,\ldots,d\}$, we define the perturbation level $\lambda(S)$ as the fraction of perturbed features: $ \lambda(S) = (d-|S|)/d\in [0,1] $. To account for different perturbation intensities, we partition $[0,1]$ into $B=10$ equal-width bins and learn a specific temperature for each bin. Let $\mathcal{B}_b := [\frac{b-1}{B}, \frac{b}{B}]$, $b \in {1,\ldots,B}$ denote these bins. Given a validation set $\mathcal{D}_{\text{val}} = {(x_i, y_i)}_{i=1}^N$, we optimize a temperature $T_b$ for each bin by minimizing the cross-entropy loss $\mathcal{L}_{\textit{CE}}$ on perturbed samples with corresponding perturbation levels:
\begin{algorithm}[H] 
\caption{$\text{ReCalX}_{\textit{TS}}$} 
\begin{algorithmic}[1] 
\Require Model $f$, validation data $\mathcal{D}_{\text{val}}$, number of bins $B$, samples per bin $M$ 
\State Initialize bins $\mathcal{B}_b = [\frac{b-1}{B}, \frac{b}{B}]$ for $b \in {1,\ldots,B}$ and Temperature parameters $\{T_b = 1\}_{b=1}^B $ 
\For{$b = 1$ to $B$} \State $\mathcal{L}_{b} = 0$ 
\Comment{Initialize loss for bin $b$} 
\For{$(x_i, y_i)$ in $\mathcal{D}_{\text{val}}$} \For{$j = 1$ to $M$} 
\State Sample $S \subset \{1, \dots, d\}$ with $\lambda(S) \in \mathcal{B}_b$ 
\Comment{Random subset with desired pert.level} 
\State $\mathcal{L}_{b} \mathrel{+}= \mathcal{L}_{\text{CE}}(f_{S}^{\pi}(x_i, T_b), y_i)$ 
\EndFor 
\EndFor 
\State $T_b^* = \argmin_{T_b>0} \mathcal{L}_b$ 
\Comment{Optimize using L-BFGS} 
\EndFor

\Return $\{T_b^*\}_{b=1}^B$ 
\end{algorithmic} 
\end{algorithm}
The algorithm iterates over each bin, sampling $M$ random subsets with appropriate perturbation levels for each validation sample. For each bin, we optimize a temperature parameter using L-BFGS \citep{liu1989limited} to minimize the cross-entropy loss on the perturbed samples. Note that in this way, we directly optimize the KL-Divergence based calibration error, while the discriminative power and ranking of predictions remains unaffected by temperature scaling. This results in a set of temperature parameters that adapt to different perturbation intensities, enabling better calibration across all explanation-relevant perturbations.
During the explanation process, ReCalX infers the perturbation-level $T_b(S)$ and applies the corresponding temperature when making predictions: 
\begin{align*} 
f_{\text{ReCalX}_{\textit{TS}}}^{\pi}(x, S; \{T_b^*\}_{b=1}^B)_k = \frac{\exp(z_k(\pi(x,S))/T_b(S))}{\sum_{j=1}^K \exp(z_j(\pi(x,S))/T_b(S))} \qquad \forall k = 1, \dots, K
\end{align*} where $b$ is the index of the bin containing the perturbation level of $S$. 
This adaptive temperature scaling ensures appropriate confidence calibration for different perturbation intensity, ultimately aiming towards more reliable explanations by reducing the maximum calibration error across all perturbation subsets $CE_{\textit{KL}}^{\max_S}$.

\section{Experiments} We evaluate ReCalX on the ImageNet \texttt{ILSVRC2012} dataset, focusing on two popular perturbation-based explanation methods, Shapley Values and LIME based on their implementation in \texttt{Captum} \citep{kokhlikyan2020captum}. For both methods, we use zero-value baseline replacement as the perturbation function $\pi$. We analyze three different architectures: DenseNet \citep{huang2017densely}, Vision Transformer (ViT) \citep{dosovitskiy2020image}, and a SigLip \citep{zhai2023sigmoid} zero-shot vision-language classier, representing diverse approaches to image classification. Pretrained models are obtained from \texttt{timm} \citep{rw2019timm} and \texttt{openclip} \citep{ilharco_gabriel_2021_5143773}  

\subsection{ReCalX reduces the Calibration Error under Perturbations}
Our first experiment evaluates whether $\text{ReCalX}_{TS}$ effectively reduces the calibration errors under subset perturbations. For this purpose we selected 200 instances from the \texttt{ImageNet} validation dataset and used $M=10$ perturbation samples to fit $\text{ReCalX}_{TS}$. Figure 1 shows the calibration error as a function of perturbation strength $\lambda_S = (d-|S|)/d$ for different architectures, both before and after applying our method each estimated using 5000 fresh instances. We explicitly evaluated the KL-Divergence based calibration error using the estimator proposed by \citep{popordanoska2024consistent} to be fully aligned with our theoretical analysis above.
For uncalibrated models, the calibration error tends to increase with higher perturbation strength across all architectures. This confirms our hypothesis that models become less reliable when processing perturbed inputs used for explanations. $\text{ReCalX}_{TS}$ substantially reduces the maximal calibration errors across all perturbation levels and architectures. The maximum calibration error drops to 0.014 for DenseNet (96\% $\downarrow$), 0.013 for ViT (95\% $\downarrow$), and 0.030 for SigLip (85\% $\downarrow$). Notably, the calibration improvement is consistent across all perturbation strengths, demonstrating that our bin-wise temperature scaling effectively adapts to different levels of feature ablation.

\begin{figure}
    \centering
    \includegraphics[width=0.99\textwidth]{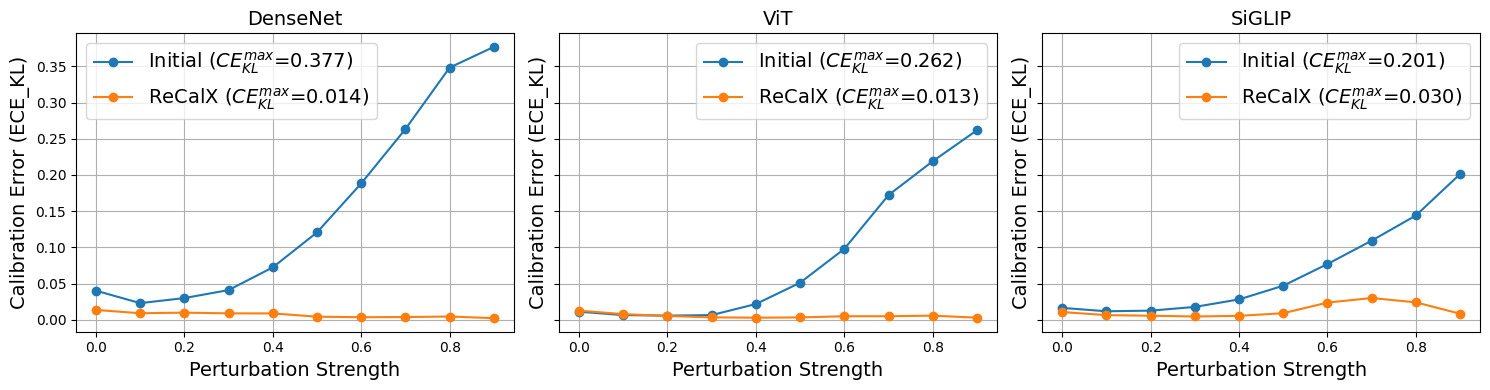}
    \caption{Evolution of calibration errors with increasing perturbation strengths (share of replaced pixels with zeros) for the considered models. The more pixels are affected by the perturbation, the error tends to increase for all uncalibrated models implying less reliable predictions. Notably, $\text{ReCalX}_\textit{TS}$ significantly reduces the calibration error across all perturbation levels and models. }
    \label{fig:enter-label}
\end{figure}

These results validate the effectiveness of $\text{ReCalX}_{TS}$ in enabling reliable predictions under the perturbations used for explanation generation. Next, we validate how this influences explanation quality.

\subsection{ReCalX improves Explanation quality}
Remember that our theoretical analysis suggests that better calibration under perturbations leads to explanations that more accurately reflect the mutual information of perturbed predictions and true labels. Thus, features that carry more information about the true class and are also utilized by the model should consequently be more relevant after calibration. To validate this empirically, we evaluate two complementary metrics, that can be considered as proxies for assessing whether explanations highlight truly informative features. (1) Human alignment \citep{fel2022harmonizing, decker2023does} measures how well explanations correspond to regions that humans consider important for classification and (2) object localization \citep{hedstrom2023quantus} quantifies how accurately explanations match the actual position of the labeled objects. For each experiment, we randomly sample 200 images from the ImageNet validation set. 

\paragraph{Human Alignment} Following \citep{fel2022harmonizing}, we measure the Spearman rank correlation between explanation importance scores and human attention annotations provided by \citep{linsley2018learning}. As shown in Table \ref{tab:results}, $\text{ReCalX}_{\textit{TS}}$ consistently improves human alignment across all models and both explanation methods. The improvements are particularly pronounced for Shapley Values, where DenseNet exhibts a 46\%, ViT a 37\% and SigLip a 13\% relative increase.
\begin{table}[h]
    \centering
    \begin{tabular}{lcccccc}
        \toprule
        & \multicolumn{2}{c}{Shapley Values} & \multicolumn{2}{c}{LIME} \\
        \cmidrule(lr){2-3} \cmidrule(lr){4-5}
        Model & Initial & $\text{ReCalX}_\textit{TS}$ & Initial & $\text{ReCalX}_\textit{TS}$ \\
        \midrule
        Densenet & 0.164 & \textbf{0.240} & 0.193 & \textbf{0.210} \\
        ViT & 0.126 & \textbf{0.172} & 0.157 & \textbf{0.168} \\
        SigLip & 0.196 & \textbf{0.221} & 0.204 & \textbf{0.211} \\
        \bottomrule
    \end{tabular}
    \caption{Human Alignment results measured via Spearman rank correlation \citep{fel2022harmonizing} for uncalibrated and calibrated models using $\text{ReCalX}_\textit{TS}$. Explanations after recalibration are consistently more aligned with Human perception for all models and methods.}
    \label{tab:results}
\end{table}
\paragraph{Object Localization} We additionally evaluate how well explanations align with actual object positions using the bounding box annotations available for ImageNet. The localization metric measures the proportion of positive importance scores that fall within the ground truth bounding box. Table \ref{tab:localization} demonstrates that recalibration with $\text{ReCalX}_{\textit{TS}}$ also leads to better object localization across all scenarios. 
\begin{table}[h]
\centering
    \begin{tabular}{lcccc}
        \toprule
        & \multicolumn{2}{c}{Shapley Values} & \multicolumn{2}{c}{LIME} \\
        \cmidrule(lr){2-3} \cmidrule(lr){4-5}
        Model & Initial & $\text{ReCalX}_\textit{TS}$ & Initial & $\text{ReCalX}_\textit{TS}$ \\
        \midrule
        Densenet & 0.3619 & \textbf{0.3816} & 0.3328 & \textbf{0.3415} \\
        ViT & 0.3345 & \textbf{0.3485} & 0.3125 & \textbf{0.3156} \\
        SigLip & 0.4548 & \textbf{0.4788} & 0.4125 & \textbf{0.4174} \\
        \bottomrule
    \end{tabular}
    \caption{Localization Metrics for different models and explanation methods. After recalibration with $\text{ReCalX}_\textit{TS}$ all models produce explanations that are more in line with the actual object position.}
       \label{tab:localization}
\end{table}

The improvements in both human alignment and localization suggest that better calibration leads to more semantically meaningful explanations, which also validate our theoretical analysis in Section 3.

\section{Discussion and Conclusion}

In this work, we established a fundamental connection between uncertainty calibration and the reliability of perturbation-based explanations. Our theoretical analysis reveals that poor calibration under feature perturbations directly impacts explanation quality, leading to potentially misleading interpretations. To address this, we introduced ReCalX, a novel recalibration approach that specifically targets explanation-relevant perturbations. Our empirical results demonstrate that ReCalX substantially reduces calibration error across different perturbation levels and consistently improves explanation quality, as measured by both human alignment and object localization metrics.
Future work could investigate if ReCalX benefits from more advanced calibration strategies beyond temperature scaling and explore other data modalities such as natural language or tabular data.

\bibliography{iclr2025_conference}
\bibliographystyle{iclr2025_conference}

\end{document}